\documentclass[twoside,journal]{IEEEtran}
\IEEEoverridecommandlockouts
\usepackage{cite}
\usepackage{amsmath,amssymb,amsfonts}
\usepackage{graphicx}
\usepackage{textcomp}
\usepackage{xcolor}
\usepackage{soul} 
\usepackage[top=0.9in, bottom=0.8in, left=0.75in, right=0.75in]{geometry}
\usepackage{amsmath} 
\usepackage{amsfonts} 
\usepackage{amssymb} 
\usepackage{amsthm}
\usepackage{algorithm} 
\usepackage{algpseudocode} 
\usepackage{graphicx} 
\usepackage{bm}
\usepackage{lipsum} 
\usepackage{listings} 
\usepackage[utf8]{inputenc} 
\usepackage[T1]{fontenc} 
\usepackage{multirow}
\usepackage[hidelinks]{hyperref}
\usepackage{breakurl}
\usepackage{booktabs}
\usepackage{xfrac}

\theoremstyle{definition}

\def\BibTeX{{\rm B\kern-.05em{\sc i\kern-.025em b}\kern-.08em
    T\kern-.1667em\lower.7ex\hbox{E}\kern-.125emX}}
\begin{document}

\title{\LARGE \textbf{ASMA: An \underline{A}daptive \underline{S}afety \underline{M}argin \underline{A}lgorithm for Vision-Language Drone Navigation via Scene-Aware Control Barrier Functions}}

\author{Sourav Sanyal, \emph{Graduate Student Member, IEEE} and Kaushik Roy, \emph{Fellow, IEEE}
\thanks{Manuscript received: March 10, 2025; Revised: June 5, 2025; Accepted: July 17, 2025.} 
\thanks{This paper was recommended for publication by Editor Giuseppe Loianno based on the evaluation of the Associate Editor and Reviewers’ comments.} 
\thanks{This work was supported by the Center for the Co-Design of Cognitive Systems (CoCoSys), a center in JUMP 2.0, an SRC program sponsored by DARPA.} 
\thanks{The authors are with School of Electrical and Computer Engineering, Purdue University, West Lafayette, IN, USA. {\tt\footnotesize \{sanyals@purdue.edu, kaushik@purdue.edu\}}} 
\thanks{Demo code available at \href{https://github.com/souravsanyal06/ASMA}{https://github.com/souravsanyal06/ASMA}} 
\thanks{Digital Object Identifier (DOI): see top of this page.}
}

\maketitle

\markboth{IEEE ROBOTICS AND AUTOMATION LETTERS. PREPRINT VERSION. ACCEPTED JULY, 2025}
{SANYAL \MakeLowercase{\textit{et al}.}: ASMA: An Adaptive Safety Margin Algorithm for Vision-Language Drone Navigation via Scene-Aware CBF}

\begin{abstract}

In the rapidly evolving field of vision–language navigation (VLN), ensuring safety for physical agents remains an open challenge. For a human-in-the-loop language-operated drone to navigate safely, it must understand natural language commands, perceive the environment, and simultaneously avoid hazards in real time. Control Barrier Functions (CBFs) are formal methods that enforce safe operating conditions. Model Predictive Control (MPC) is an optimization framework that plans a sequence of future actions over a prediction horizon, ensuring smooth trajectory tracking while obeying constraints. In this work, we consider a VLN-operated drone platform and enhance its safety by formulating a novel scene-aware CBF that leverages ego-centric observations from a camera which has both Red-Green-Blue as well as Depth (RGB-D) channels. A CBF-less baseline system uses a Vision–Language Encoder with cross–modal attention to convert commands into an ordered sequence of landmarks. An object detection model identifies and verifies these landmarks in the captured images to generate a planned path. To further enhance safety, an Adaptive Safety Margin Algorithm (ASMA) is proposed. ASMA tracks moving objects and performs scene-aware CBF evaluation on-the-fly, which serves as an additional constraint within the MPC framework. By continuously identifying potentially risky observations, the system performs prediction in real time about unsafe conditions and proactively adjusts its control actions to maintain safe navigation throughout the trajectory. Deployed on a Parrot Bebop2 quadrotor in the Gazebo environment using the Robot Operating System (ROS), ASMA achieves 64\%–67\% increase in success rates with only a slight  increase (1.4\%–5.8\%) in trajectory lengths compared to the baseline CBF-less VLN.

\end{abstract}


\section{Introduction}

\IEEEPARstart{F}oundational models pretrained on exa-scale internet data have made significant strides in vision and language processing tasks with little to no fine-tuning, as exemplified by a new family of AI models such as BERT~\cite{devlin2018bert}, GPT-3~\cite{NEURIPS2020_1457c0d6}, GPT-4~\cite{achiam2023gpt}, CLIP~\cite{radford2021learning}, DALL-E~\cite{ramesh2021zero}, and PALM-E~\cite{driess2023palm}. The fusion of vision and language models~\cite{radford2021learning, driess2023palm} has enabled machines to interact with operating environments in increasingly intuitive ways. As VLN models become more widespread, the once sci-fi dream of robots understanding and interacting in complex environments through natural language commands is now a reality. This has been enabled by the emerging field of vision-language navigation (VLN)~\cite{shah2023lm, krantz2020beyond, hong2022bridging, anderson2018vision, hong2021vln, cui2023grounded}. Autonomous drones, pivotal in smart agriculture, search and rescue, and firefighting~\cite{giones2017toys}, are set to contribute up to \$54.6 billion to the global economy by 2030~\cite{droneii2024}. Imagine a scenario where VLN powered drones translate human-specified contextual instructions into actions. However, for VLNs, navigating dynamic environments using robot vision remains an open research problem. 
To that effect, Control Barrier Functions (CBFs)~\cite{ames2019control, ames2016control} provide a formal mathematical framework for enforcing safe operating conditions in dynamical systems, making them useful for real-time applications where safety is crucial. On the other hand, Model Predictive Control (MPC) is an optimization framework that plans a sequence of future actions over a prediction horizon to ensure smooth trajectory tracking while obeying constraints.

We address the challenge of safe VLN for drone navigation by introducing ASMA—an Adaptive Safety Margin Algorithm that combines high-level vision-language reasoning with low-level safety control. Starting from a baseline system with a cross-modal Vision–Language Encoder \cite{openai2021clip} and YOLOv5-based object detection \cite{yolov5}, ASMA translates language commands into ordered landmarks. To handle dynamic hazards, we propose a scene-aware CBF that uses real-time RGB–D observations for safety-aware control.
ASMA adjusts drone control by tracking moving objects, predicting safety risks, and applying scene-aware CBFs within an MPC framework. It proactively responds to unsafe conditions in real time to ensure safe navigation throughout the trajectory.
The proposed integration of VLNs with CBFs provides a formal safety layer which enhances VLN reliability of physical agents (in this work a drone).
\begin{figure*}[t]
    \centering
    \includegraphics[width=0.8\textwidth]{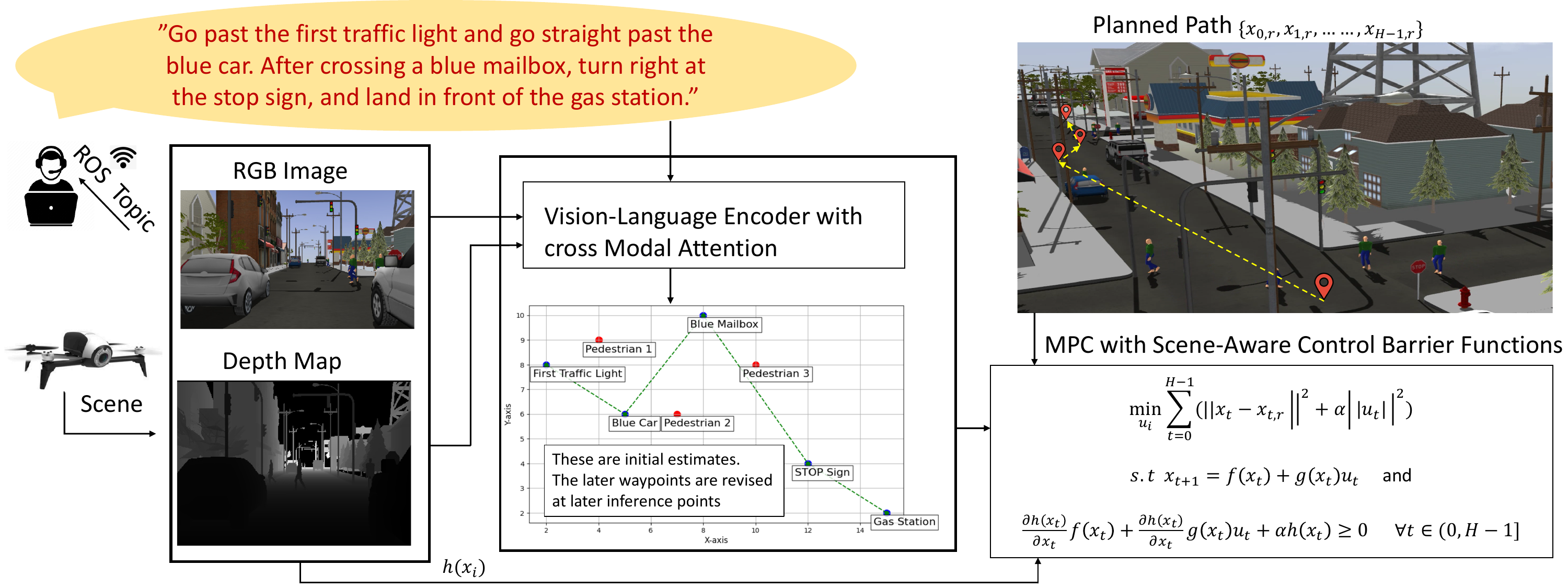}
    \caption{Overview of the proposed ASMA framework. The system takes natural language instructions and processes RGB and depth data to create a 2D map with language-grounded landmarks and dynamic obstacles. A planned path is generated, and MPC with Scene-Aware Control Barrier Functions (CBFs) ensures safe navigation along the trajectory.}
    \label{fig:asma_overview}
\end{figure*}

Our main contributions are:
\begin{itemize}
    \item A vision-language encoder with attention and detection for instruction-driven planning
 (Section~\ref{sec:crossmodal}).
    \item A scene-aware CBF for safe navigation in dynamic environments (Section~\ref{sacbf}).
    \item A full-stack modular system integrated in ROS for real-time navigation (Section~\ref{asma}).
    \item Extensive evaluation showing improved safety and accuracy over a CBF-less VLN baseline in ROS-Gazebo on a Parrot Bebop2 quadrotor (Section~\ref{res}).
\end{itemize}

In simulation, ASMA achieves  64\%--67\% increase in success rates with only a slight (1.4\%--5.8\%) increase in trajectory lengths compared to the baseline CBF-less VLN. 

\vspace{-1mm}
\section{Related Work}
\label{sec:related_work}
\noindent{\textbf{Vision-Language Models for Robot Navigation:}} In Vision-Language Navigation (VLN), agents interpret language commands to navigate through environments using visual cues \cite{shah2023lm, krantz2020beyond, hong2022bridging, anderson2018vision, hong2021vln, cui2023grounded}. Previous works, such as \cite{krantz2020beyond, hong2022bridging}, have expanded VLN into continuous environments (VLN-CE). Works in \cite{anderson2018vision, hong2021vln, cui2023grounded} have explored VLN focusing on interpreting visually-grounded instructions and developing models like VLN BERT to improve navigation performance through entity-landmark pre-training techniques. {Safe-VLN \mbox{\cite{yue2024safe}} employs 2D LiDAR for safer waypoint prediction in VLN-CE, however the work only considers static objects along the navigation path.} The works in \cite{huang2023visual, wang2023gridmm} integrate pretrained visual-language features with navigation maps. The work in \cite{lin2022adapt} utilizes action prompts for improved spatial navigation precision. Room2Room \cite{pejsa2016room2room} enables teleoperated communication using augmented reality, and \cite{an2024etpnav} introduces the ‘Tryout’ method to prevent collision-related navigational stalls. However, these approaches do not address the physical dynamics of robots, crucial for verifying safety. Our work focuses on a teleoperated drone similar to \cite{ibrahimov2019dronepick} with VLN capabilities, utilizing an RGB-D sensor {and aims to enhance its safety and reliability in presence of dynamic objects (moving arbitrarily) and complex structures.}\\

\noindent{\textbf{Control Barrier Functions for Safety:}} 
Control barrier functions (CBFs) are essential tools from robust control theory, ensuring safety constraints are maintained in dynamic systems \cite{ames2019control, ames2016control}. 
By formally defining safe boundaries, CBFs dynamically adjust control actions to prevent safety violations. {Vision-based control barrier functions (V-CBFs) \mbox{\cite{abdi2023safe}} extend these protocols using conditional generative adversarial networks (C-GANs) to infer geometric safety constraints directly from visual observations. 
Neural formulations such as BarrierNet \mbox{\cite{xiao2023barriernet}} integrate differentiable control barrier functions into end-to-end learning pipelines while maintaining formal safety guarantees.} Other related works include \cite{de2024point}, which develops a low-cost method for synthesizing quadratic CBFs over point cloud data, and \cite{sankaranarayanan2024cbf}, which addresses landing safety for aerial robots. 
{However, none of these methods incorporate language-conditioned goals or semantic scene understanding.} {In contrast, our method dynamically synthesizes scene-aware CBFs grounded in both semantic visual understanding and language-driven intent, enabling reactive and context-aware safety enforcement in the presence of dynamic obstacles and complex structures. This allows the robot to proactively adapt its trajectory even without operator's explicit intervention, based on high-level task understanding and visual semantics, which existing visual or neural CBF methods do not support.}

\vspace{1\baselineskip}

{Although there exists several datasets and benchmarks for VLN, they do not take into account environmental changes which require online re-planning. Moreover, these datasets do not evaluate the interaction between language-driven intent and dynamic safety requirements, which our work explicitly addresses. 
Furthermore, the main aim in this work is not to compete with them, but to investigate the safety aspect of VLNs by synthesizing scene-aware CBFs on the fly. To that effect, this work considers two environments built from scratch using Gazebo and is tested on a Parrot Bebop2 quadrotor by considering the necessary robot-environment dynamics required for simulating and testing the safety and reliability challenges for agentic VLNs in a physical world.}
\begin{figure*}[!t]
    \centering
    \includegraphics[width=0.8\textwidth]{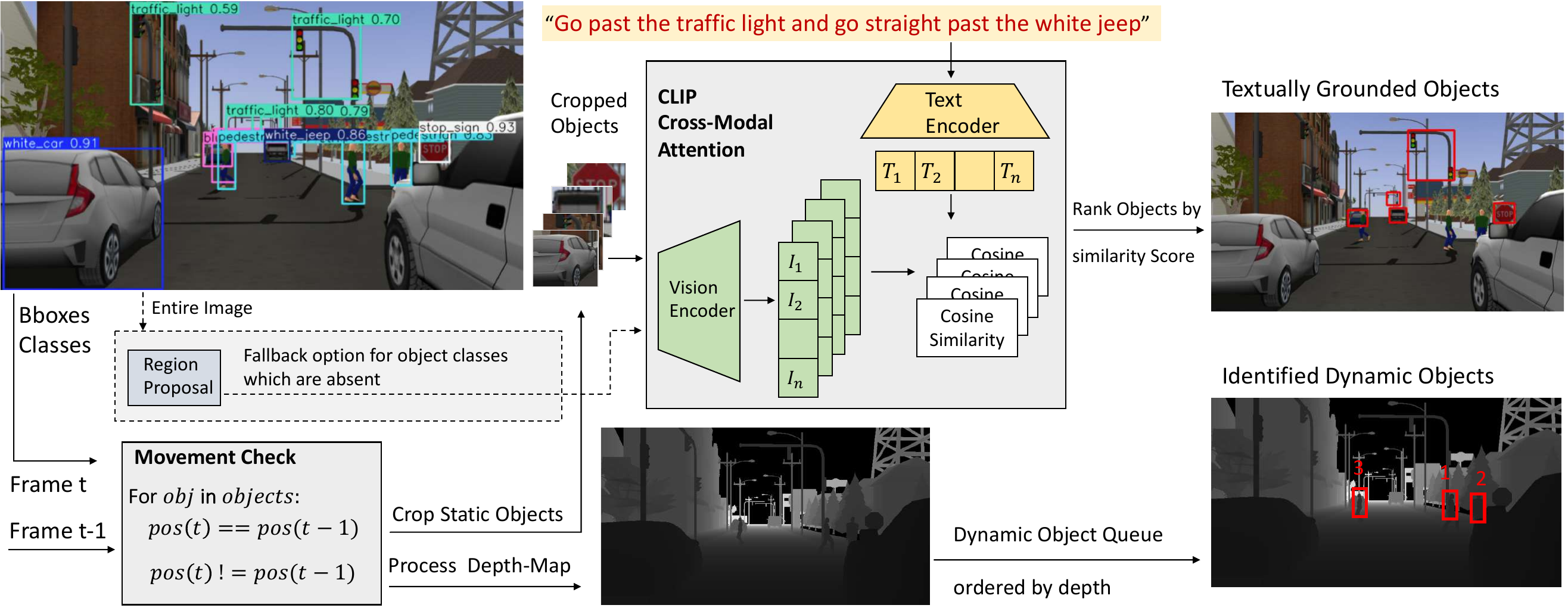}
    \caption{Illustration of the multi-modal cross-attention pipeline for text-conditioned navigation. The system first detects objects using YOLOv5, followed by a movement check to classify static and dynamic objects across frames. Static objects are cropped and fed into a CLIP-based cross-modal attention module, which ranks objects based on textual relevance. For dynamic objects, a depth map is used to sort them by distance, enabling downstream obstacle-aware planning. A fallback mechanism is included to handle objects not recognized by YOLO, using region proposals and CLIP-based zero-shot recognition.}
    \label{fig:CMA}
\end{figure*}

\section{Proposed Approach}
We deploy a Parrot Bebop 2 quadrotor in a ROS-Gazebo environment with an RGB-D sensor, following \cite{sanyal2023ramp, sanyal2024ev, joshi2024real}. We introduce ASMA—an Adaptive Safety Margin Algorithm for drone VLN (Figure~\ref{fig:asma_overview}). The system processes language instructions and RGB-D input via a Vision-Language Encoder with cross-modal attention to build a language-grounded 2D map of landmarks and obstacles. A planned path is computed, and a Scene-Aware CBF-based MPC policy ensures safe, adaptive navigation.

\subsection{\textbf{Vision-Language Encoder with Cross-Modal Attention}}
\label{sec:crossmodal}

Figure~\ref{fig:CMA} shows our pipeline for grounding language in visual objects. YOLOv5 detects objects, and CLIP matches cropped regions to instruction-referenced landmarks via a shared embedding space. Before processing, we distinguish dynamic from static objects by comparing bounding box centers over time. Let $\text{pos}_i(t)$ be the center of object $i$ at time $t$. If
\begin{equation}
\text{pos}_i(t) = \text{pos}_i(t-1),
\end{equation}
the object is static; otherwise,
\begin{equation}
\text{pos}_i(t) \neq \text{pos}_i(t-1),
\end{equation}
it is dynamic and prioritized for collision avoidance. Static objects are matched via CLIP, while dynamic ones are handled through depth maps. The depth map also acts as instance segmentation, aiding avoidance of large structures like walls and trees for broader scene awareness.

Given a natural language instruction (e.g., \emph{"Go past the first traffic light and go straight past the blue car."}), we extract an ordered list of landmark tokens such as \texttt{traffic light} and \texttt{blue car}. The onboard RGB image $\mathbf{I}$ is passed to YOLOv5, which outputs bounding boxes:
\begin{equation}
\mathbf{bbox}_i = (x_1^{(i)},\, y_1^{(i)},\, x_2^{(i)},\, y_2^{(i)},\, l_i), \quad i = 1,\ldots,N.
\end{equation}
We crop each object from the image:
\begin{equation}
\mathbf{I}_i^{\text{crop}} = \mathrm{CropImage}(\mathbf{I}, \mathbf{bbox}_i).
\end{equation}
Each landmark phrase $\mathbf{T}_\ell$ is encoded by CLIP’s text encoder:
\begin{equation}
\mathbf{z}_{\text{text}}^\ell = \Phi_{\text{text}}(\mathbf{T}_\ell),
\end{equation}
and each object crop is encoded via CLIP’s image encoder:
\begin{equation}
\mathbf{z}_{\text{obj}}^i = \Phi_{\text{img}}(\mathbf{I}_i^{\text{crop}}).
\end{equation}
Cosine similarity is computed as:
\begin{equation}
S_{\ell,i} = 
\frac{
\mathbf{z}_{\text{obj}}^i \cdot \mathbf{z}_{\text{text}}^\ell
}{
\|\mathbf{z}_{\text{obj}}^i\|\, \|\mathbf{z}_{\text{text}}^\ell\|
}.
\label{eq:landmark_similarity}
\end{equation}
{The bounding box $\mathbf{bbox}_{i}$ with the highest score is selected if $S_{\ell,i} > \theta$, where $\theta$ is a confidence threshold. If no match exceeds $\theta$, we fall back to a lightweight region proposal strategy that samples a small number ($10-15$) of image patches for CLIP-based matching. This fallback is rarely triggered in practice, as modern detectors trained on open-vocabulary datasets cover a broad range of objects. If no match is found, the landmark is flagged as unresolved (see Section\mbox{~\ref{sec:disc}}).}

\noindent \textbf{Path Generation with Landmark Order:}
Bounding-box centers \(\mathbf{p}_\ell\) for each landmark~\(\ell\) are arranged in the sequence specified by the user:
\[
\mathbf{p}_1 \;\rightarrow\; \mathbf{p}_2 \;\rightarrow\; \dots \;\rightarrow\; \mathbf{p}_L.
\]
A path planner connects these waypoints into a trajectory. As the drone moves, the list is dynamically updated as new landmarks become visible. By preserving instruction order and using CLIP similarity, the system ensures sequential, scene-aware navigation with safety guarantees.

\subsection{\textbf{Enhancing VLN with Formal Safety Methods}}
\label{sacbf}
Control Barrier Functions (CBFs) are essential tools in safety-critical control systems that enforce safety constraints through mathematical functions. In this work, we integrate CBFs within a Model Predictive Control (MPC) framework, ensuring that safety constraints are proactively enforced while the drone follows its planned trajectory.

\begin{figure}[!t]
\vspace{-3mm}
\begin{center}
   \includegraphics[width = 0.4\textwidth]{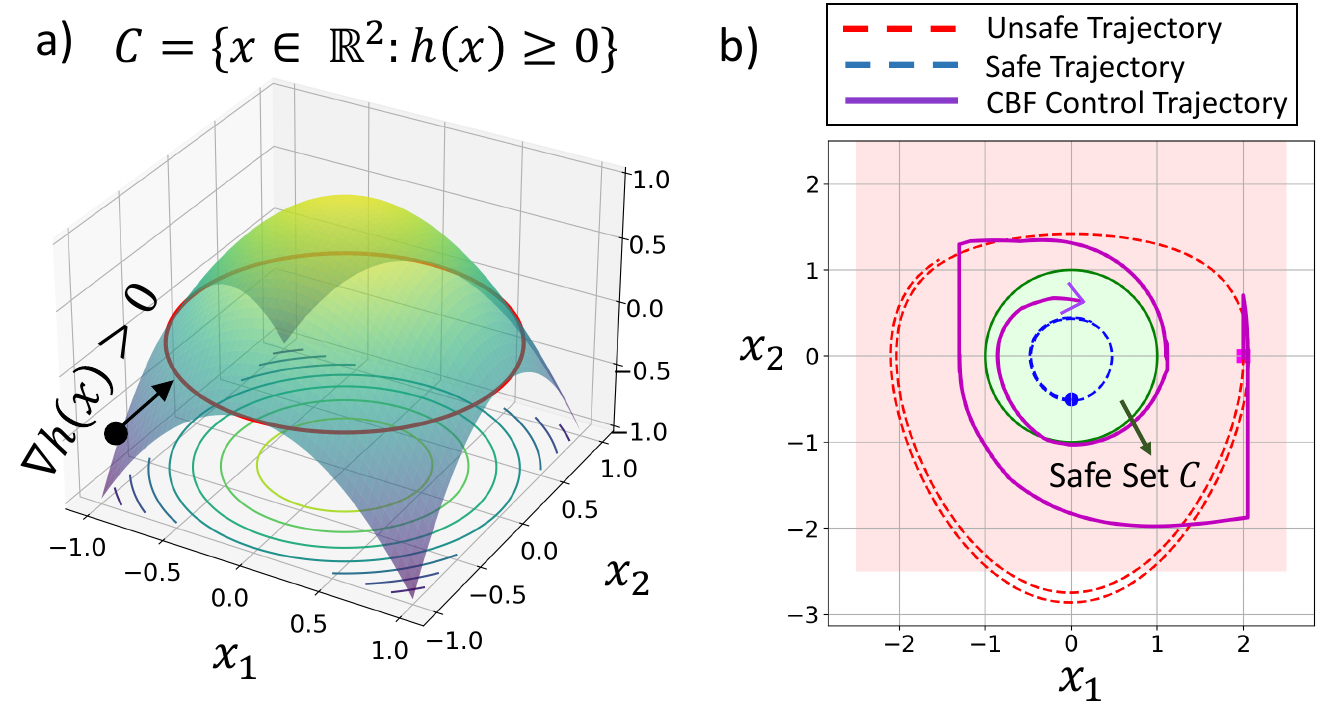} 
   \vspace{-3mm}
\caption{Toy Illustration of Control Barrier Functions. (a) 3D view of the safe set $\mathcal{C}$ where $h(\bm{x}) \geq 0$ (above red ring). (b) Comparison of trajectories: unsafe (red dashed), safe (blue dashed), and CBF-controlled (magenta solid).
{The CBF-controlled trajectory starts outside the safe set and is driven into the safe region by the adaptive controller, demonstrating the forward-invariance property once inside.}}
   \label{cbf_fig}
\end{center}
\vspace{-4mm}
\end{figure}

\subsubsection{\textbf{Preliminaries}}
We consider a quadrotor described by the following non-linear control affine dynamics with an ego-centric depth-map:

\begin{equation}
    \dot{\bm{x}} = f(\bm{x}) + g(\bm{x})\bm{u}, \quad \boldsymbol{\xi} = \Psi({\bm{x}}, \bm{d})
    \label{system}
\end{equation}

Here, \( \dot{\bm{x}} \) denotes the time derivative of the state vector \( \bm{x} \in \mathbb{R}^{12} \), covering the drone's positions, orientations, and velocities. {The control input \mbox{\( \bm{u} \in \mathbb{R}^3 \)} represents the commanded linear velocity of the quadrotor. This velocity-level abstraction assumes that the onboard flight controller is responsible for low-level stabilization using thrust and attitude control. This allows us to define first-order CBFs (later in Section \mbox{\ref{subsec:sacbf})} with respect to the velocity-controlled dynamics.}
 Functions \( f(\bm{x}) \) form the autonomous dynamics and \( g(\bm{x}) \) signifies the dynamics that can be controlled in an affine manner. Additionally, the depth-map \( \bm{d} \) provides obstacle distances, with \( \bm{\xi} = \Psi(\bm{x}, \bm{d}) \) transforming these distances from pixel to physical space, incorporating the drone's current position. Let \(\mathcal{C} \subset \mathbb{R}^{12}\) represent a safety set defined through a continuously differentiable function \(h(\bm{x})\) such that:
\begin{equation}
    \mathcal{C} = \{\bm{x} \in \mathbb{R}^{12} : h(\bm{x}) \geq 0\}
\end{equation}

Figure~\ref{cbf_fig}a illustrates a toy safety set \(\mathcal{C}\), marked by the boundary where \(h(\bm{x}) = 0\) and the region where \(h(\bm{x}) > 0\), indicating safe operational zones. The function \(h(\bm{x})\) is characterized by its lie derivatives:
\begin{subequations}
\begin{align}
    L_f h(\bm{x}) &= \nabla h(\bm{x}) \cdot f(\bm{x}), \\
    L_g h(\bm{x}) &= \nabla h(\bm{x}) \cdot g(\bm{x}),
\end{align}
\end{subequations}
which are critical for monitoring the system’s safety relative to state changes and control action changes. 

\textbf{Theorem (Safety Verification):} For safety verification, it is required that:
\begin{equation}
    L_f h(\bm{x}) + L_g h(\bm{x})\bm{u} + \alpha(h(\bm{x})) \geq 0,
    \label{constraint}
\end{equation}
where \(\alpha\) is a class \(\mathcal{K}\) function (meaning $\alpha(0) = 0$ and $\alpha(\mathcal{K}x_2) > \alpha(\mathcal{K}x_1)$ $\forall$ $x_2>x_1$ and $\mathcal{K} >0 $).  $\nabla h(\bm{x}) > $ 0 in the unsafe region ($h(\bm{x}) < $ 0) will drive $h(\bm{x})$ to become positive again.

\textbf{Corollary (CBF-MPC Integration):} \emph{If a function $h(\bm{x})$ can be designed such that adjustments in $\bm{u}$ satisfy the constraint in Eqn.~(\ref{constraint}) at every time step, then embedding these constraints inside an MPC framework ensures that the system guarantees safety while optimizing trajectory tracking.}

This forms the core of our Scene-Aware CBF-MPC framework (Algorithm 1), where the safety constraint is included as part of the trajectory optimization problem, rather than being enforced in a purely reactive manner.

\begin{equation}
    \min_{\bm{u}_0, \dots, \bm{u}_{H-1}} J = \sum_{t=0}^{H-1} \Bigl\{
    \|\bm{x}_t - \bm{x}_{r,t}\|^2 + \alpha \|\bm{u}_t\|^2
    \Bigr\},
    \label{mpc_cost}
\end{equation}
subject to:
\begin{equation}
    \bm{x}_{t+1} = f(\bm{x}_t) + g(\bm{x}_t) \bm{u}_t, \quad \forall t \in (0,H-1]
\end{equation}
\begin{equation}
    L_f h(\bm{x}_t) + L_g h(\bm{x}_t) \bm{u}_t + \alpha h(\bm{x}_t) \geq 0, \quad \forall t  \in (0,H-1]
\end{equation}
\begin{equation}
   \implies  
   \begin{aligned}
       & \underbrace{\frac{\partial h(\bm{x}_t)}{\partial \bm{x}_t} f(\bm{x}_t) + 
       \frac{\partial h(\bm{x}_t)}{\partial \bm{x}_t} g(\bm{x}_t) \bm{u}_t }_{\text{Expanded form of Lie derivatives}} \\
       & \quad + \alpha h(\bm{x}_t) \geq 0,  \quad \forall t  \in (0,H-1]
   \end{aligned}
   \label{mpc_cbf}
\end{equation}

where \( \bm{x}_{r,t} \) denotes the reference trajectory computed from waypoints extracted by the vision-language encoder.

\begin{algorithm}[!t]
\small
\caption{Scene-Aware CBF-MPC}
\label{alg:solve_qp}
\begin{algorithmic}[1]
\Function{SA-CBF-MPC}{$\mathbf{p_{curr}}, \mathbf{X}_{\text{plan}}, h(\bm{x}), \text{Obstacles}$}
    \State Formulate MPC optimization problem over horizon \( H \)
    \For{each time step } $t = 0$ \textbf{to} $H-1$
        \State Extract reference state $\mathbf{x}_{r,t}$ from $\mathbf{X}_{\text{plan}}$
        \For{each obstacle in Obstacles}
            \State Compute \( h(\bm{x})\) using Equation \ref{h_x}
            \If{\( h(\bm{x}) \) < 0}
                \State Apply CBF constraints to enforce safety
            \EndIf
        \EndFor
    \EndFor
    \State Solve MPC optimization  incorporating CBF constraints
    \State \Return optimal control sequence \( \bm{U} = [\bm{u}_0, \bm{u}_1, ..., \bm{u}_{H-1}] \)
\EndFunction
\end{algorithmic}
\vspace{-0.4mm}
\end{algorithm}

To integrate vision-language and depth outputs into the control loop, we convert pixel-level detections to 3D global positions. For each pixel \((i,j)\) in the cropped depth map with depth \(d\), we use intrinsic parameters—focal length \(f\) and camera center \((c_x, c_y)\)—to compute 3D coordinates in the camera frame:

\begin{equation}
X_c = (i - c_x) \cdot \frac{d}{f}, 
\quad
Y_c = (j - c_y) \cdot \frac{d}{f}, 
\quad
Z_c = d.
\end{equation}
These coordinates are then transformed into the global frame:
\begin{equation}
\begin{bmatrix}
X_g\\
Y_g\\
Z_g
\end{bmatrix}
=
\mathbf{R}
\cdot
\begin{bmatrix}
X_c\\
Y_c\\
Z_c
\end{bmatrix}
+
\vec{\mathbf{p}}_{\text{curr}},
\end{equation}
Here, \(\mathbf{R}\) is the rotation matrix for the drone’s heading, and \(\vec{\mathbf{p}}_{\text{curr}}\) is its estimated global position. Both are updated in real time using an EKF~\cite{fernandes2023gnss} to reduce sensor drift. The vision-language module detects targets and obstacles via bounding boxes; for each box center \((i,j)\), we extract depth \(d\) and compute global coordinates \(\vec{\mathbf{p}}_{\text{target}}\) or \(\vec{\mathbf{p}}_{\text{obs}}\). Let
\begin{equation}
\vec{\mathbf{d}}_{\text{target}} = \vec{\mathbf{p}}_{\text{target}} - \vec{\mathbf{p}}_{\text{curr}}, \quad
\vec{\mathbf{d}}_{\text{obs}} = \vec{\mathbf{p}}_{\text{obs}} - \vec{\mathbf{p}}_{\text{curr}}.
\label{eq:direction_vectors}
\end{equation}
Obstacles are prioritized by
\begin{equation}
\text{Priority} = \frac{1}{\|\vec{\mathbf{d}}_{\text{obs}}\|},
\label{eq:priority}
\end{equation}
so nearer obstacles are processed first. If a high-priority obstacle lies along the path, we compute
\begin{equation}
\sigma_d = \text{sign}\left((\vec{\mathbf{d}}_{\text{target}} \times \vec{\mathbf{d}}_{\text{obs}})_z\right),
\label{eq:sigma_d}
\end{equation}
which determines whether the obstacle lies to the left (\(\sigma_d < 0\)) or right (\(\sigma_d > 0\)) of the target direction. This is used in Equation~(23a) to steer appropriately and avoid the obstacle.

\subsubsection{\textbf{Scene-Aware CBF}}
\label{subsec:sacbf}
{Since we operate at the velocity-control level, where control inputs correspond to desired velocities (Twist messages), we formulate first-order CBFs with respect to this abstraction.}
The proposed Scene-Aware CBF which is used within the MPC framework is defined as:
\begin{equation}
h(\bm{x}) = 
\begin{bmatrix}
\vec{\mathbf{d}}_{\text{obs}}  - \mathbf{d_{\text{safe}}} \\
\theta - \theta_{\text{safe}}
\end{bmatrix}
\label{h_x}
\end{equation}
\begin{equation}
\theta 
= 
\cos^{-1}\!\Bigl(
\frac{
\vec{\mathbf{d}}_{\text{target}} \cdot \vec{\mathbf{d}}_{\text{obs}}
}{
\|\vec{\mathbf{d}}_{\text{target}}\|\,
\|\vec{\mathbf{d}}_{\text{obs}}\|
}
\Bigr),
\label{theta}
\end{equation}
where \(\theta\) is the angle between \(\vec{\mathbf{d}}_{\text{obs}}\) and \(\vec{\mathbf{d}}_{\text{target}}\). The gradient for each component of \(h(\bm{x})\)is given by:
\begin{subequations}
    \begin{align}
\frac{\partial h_1(\bm{x})}{\partial \bm{x}} &= \sigma_{d} \,\frac{\partial \vec{\mathbf{d}}_{\text{obs}}}{\partial \bm{x}}, \\
\frac{\partial h_2(\bm{x})}{\partial \bm{x}} &= -\,\csc\theta \,\frac{\partial \cos\theta}{\partial \bm{x}}.
\end{align}
\label{delta_h}
\end{subequations}

These gradients feed into the Lie derivatives in Eq.~(\ref{constraint}), allowing the optimizer to adjust control inputs for safety. When multiple obstacles are detected, a priority queue processes them by increasing \(\|\vec{\mathbf{d}}_{\text{obs}}\|\), ensuring the closest ones are handled first.

\begin{algorithm}[!t]
\small
\caption{Adaptive Safety Margin Algorithm (ASMA)}
\label{alg:asma}
\begin{algorithmic}[1]
\Require RGB-D image $\mathbf{I}$, VLN instruction $\mathbf{cmd}$
\Ensure Drone command vector $\mathbf{u}$ with safety constraints

\State Initialize $\mathbf{global\_image}$, $\mathbf{global\_depth}$ from RGB-D
\While{not \Call{rospy.is\_shutdown()}{}}

    \State $\mathbf{L} \leftarrow \mathbf{ExtractLandmarks}(\mathbf{cmd})$ \Comment{Tokenize instruction}

    \State \textbf{Start Thread A}:
    \State Acquire $\mathbf{image\_lock}$ \Comment{Wait for $\mathbf{I}$}
        \State \hspace{4mm} $\mathbf{bbox} \leftarrow \mathbf{DetectRelevantObjects}(\mathbf{global\_img}, \mathbf{L})$
        \State \hspace{4mm} $\mathbf{I_{crop}} \leftarrow \mathbf{CropObjects}(\mathbf{global\_img}, \mathbf{bbox})$
        \State \hspace{4mm} $\mathbf{D_{crop}} \leftarrow \mathbf{CropDepth}(\mathbf{global\_depth}, \mathbf{bbox})$
    \State Release $\mathbf{image\_lock}$

    \State \textbf{Start Thread B}:
    \State Acquire $\mathbf{image\_lock}$ \Comment{Wait for $\mathbf{I}$ and cropped objects}
        \State \hspace{4mm} $\mathbf{E_{\text{objs}}} \leftarrow \mathbf{\Phi_{img}(I_{crop})}$
        \State \hspace{4mm} $\mathbf{E_{\text{text}}} \leftarrow \mathbf{\Phi_{text}(L)}$
        \State \hspace{4mm} $\mathbf{S} \leftarrow \mathbf{ComputeSimilarity}(\mathbf{E_{\text{objs}}}, \mathbf{E_{\text{text}}})$
    \State Release $\mathbf{image\_lock}$

    \State $\mathbf{X}_{\text{plan}} \leftarrow \mathbf{GeneratePath}(\mathbf{L}, \mathbf{bbox}, \mathbf{D_{crop}})$

\If{any $\mathbf{L}$ not in detected object classes}
    \State $\mathbf{bbox}_{\text{fallback}} \leftarrow \mathbf{RegionProposal}(\mathbf{global\_img})$
    \State $\mathbf{X}_{\text{plan}} \leftarrow \mathbf{UpdatePath}(\mathbf{X}_{\text{plan}}, \mathbf{bbox}_{\text{fallback}})$
\EndIf

    \State $\bm{u} \leftarrow \mathbf{SA-CBF-MPC}(\vec{\mathbf{p}}_{\text{curr}}, \mathbf{X}_{\text{plan}}, h(\bm{x}), \{\vec{\mathbf{p}}_{\text{obs}}\})$

    \State \text{PUBLISH\_CONTROL}($\bm{u}$)

\EndWhile

\end{algorithmic}
\end{algorithm}

\vspace{-3mm}
\subsection{\textbf{Adaptive Safety Margin Algorithm (ASMA)}}
\label{asma}
Algorithm~\ref{alg:asma} gives a walkthrough of the entire system. Each iteration begins by tokenizing the VLN instruction (line~3). Thread~A locks the image buffer (line~5), detects objects (line~6), and crops RGB and depth data (lines~7--8). Meanwhile, Thread~B computes similarity scores between the cropped objects and instruction tokens (lines~12--15). A planning module then generates a trajectory using these detections (line~17), invoking a fallback if landmarks are missing (lines~18--20). Finally, the planned trajectory is passed to the SA-CBF-MPC module (line~21), which enforces safety and publishes the control command (line~22). This process repeats continuously.

\section{Results}
\label{res}
\subsection{\textbf{Methodology}}
We implemented the ASMA framework in ROS on a parrot bebop2 quadrotor within the Gazebo environment. The pretrained CLIP model was obtained from OpenAI's repository \cite{openai2021clip}. Object detection was performed using YOLOv5 \cite{yolov5} ($\sim$21 million parameters), with training data annotated via LabelImg \cite{labelimg2020}. The pretrained CLIP consisted of $\sim$149 million parameters. Because of the large model sizes, we performed thread synchronization (Algorithm 2) to toggle between the two inference modes. The threshold $\theta$ in Vision-Language Encoder was set to 0.2. Scene-Aware CBF with MPC optimizations sampled at 5 Hz (to give sufficient time to the Vision-Language Encoder) were conducted using cvxopt \cite{cvxopt}, and the RotorS simulator \cite{rotorS} was used to integrate lower level control. $\vec{\mathbf{d}}_{\textbf{safe}}$ was set to 2 meters and $ \theta_{\textbf{safe}}$ to 30 degrees. RGB-D focal length $f$ was 10 meters.
\vspace{-2mm}

\subsection{\textbf{Comparative Schemes}}
\label{subsec: comp}
\begin{itemize}
\item {\textbf{CBF-less VLN:} A baseline vision-language navigation (VLN) method that uses CLIP-YOLO for grounding instructions and a basic flight controller to execute trajectories, but does not enforce safety using control barrier functions (CBFs). This captures how standard VLN systems operate without dynamic scene-aware safety.}

\item {\textbf{ASMA-Reactive:} Implements scene-aware CBFs based on the constraint in Equation~\mbox{(\ref{constraint})}, enforcing safety reactively. The nominal control here is a basic PID. Forward invariance is restored upon violation by driving the system back into the safe set.}

\item {\textbf{ASMA-MPC:} Integrates CBF constraints within an MPC framework for pro-active safety enforcement. Forward invariance is proactively maintained by forecasting safety violations over the prediction horizon and adjusting the trajectory accordingly.}

\item {\textbf{CBF-Only Navigation:} A language-agnostic baseline that uses CBFs for collision avoidance, assuming perfect knowledge of obstacle poses from simulation. It navigates toward a fixed target without any vision-language grounding, serving as a safety upper bound without perception or instruction-following.}
\end{itemize}

\begin{figure}[t]
    \centering
    \includegraphics[width=0.83\columnwidth]{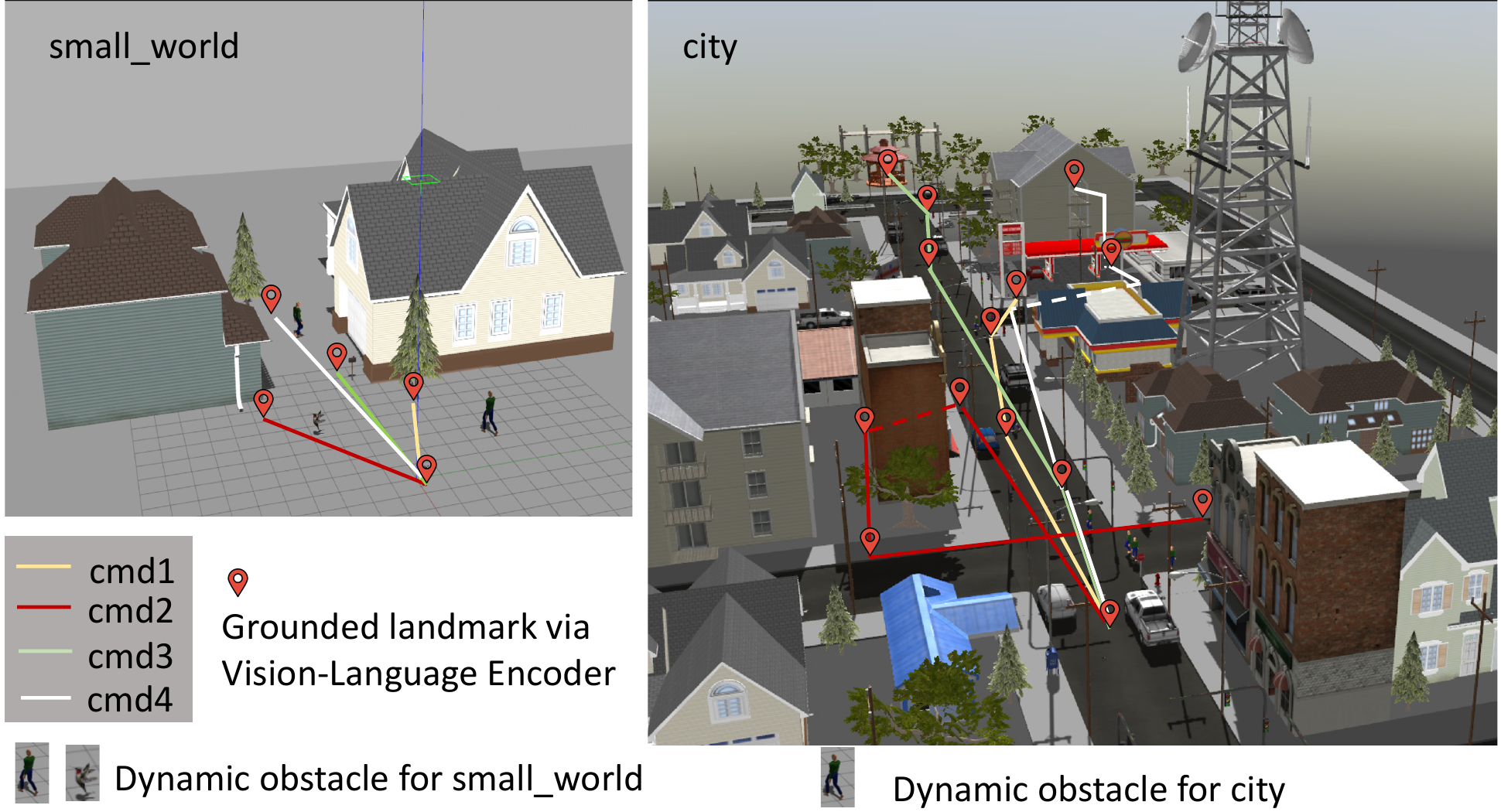}
    \caption{Gazebo environments -- \textit{\textbf{left:} small\_world}, \textbf{right:} \textit{city}}
    \label{fig:envs}
\end{figure}

\begin{table}[t]
\renewcommand{\arraystretch}{0.5}
    \centering
    \caption{Vision--Language Encoder Performance.}
    \label{tab:vl_performance}
    \begin{tabular}{lcc}
        \toprule
        Environment & Average Similarity Score & Grounding Accuracy (\%) \\
        \midrule
        small\_world & 0.36 & 98 \\
        city         & 0.31 & 92 \\
        \bottomrule
    \end{tabular}
\end{table}

We test our method in two simulated environments (Figure \ref{fig:envs}): the \textit{small world} for the CBF-based MPC formulation and the more complex and cluttered \textit{city} environment for planning  (Table~\ref{tab:vln}). {In \textit{city}, commands include flying through a narrow alley (\textbf{cmd2}), entering a gazebo and landing (\textbf{cmd3}), and ascending to a third-floor balcony to land (\textbf{cmd4}), all requiring spatial reasoning. Some instructions also contain logical conditionals—for example, \textbf{cmd2} includes “if a white truck is visible, land in front of it,” which is grounded using vision-language attention. While the current examples do not include explicit temporal instructions (e.g., “wait for 5 seconds”), the framework supports such extensions via the same cross-modal grounding mechanism.} See the additional media file for simulations.
Performance is evaluated using Trajectory Length (TL), Success Rate (SR; trials reaching within 1 meter of the target), and Navigation Error (NE; the final Euclidean distance from the target).

\begin{table*}[!t]
\renewcommand{\arraystretch}{0.72}
\centering
\caption{VLN performance in two environments. $\downarrow$ indicates lower is better. $\uparrow$ implies higher is better. }
\label{tab:vln}
\resizebox{\textwidth}{!}{%
\begin{tabular}{l l l c c c c}
\toprule
\textbf{\small{Environment}}
& \textbf{\small{VLN Instruction}} 
& \textbf{\small{Metric}} 
& \textbf{\small{CBF-Less}}
& \textbf{\small{ASMA-Reactive}}
& \textbf{\small{ASMA-MPC}}
& \textbf{\small{CBF-Only}} \\
\midrule

\multirow{12}{*}{\textbf{\small{small\_world}}}

& \multirow{3}{*}{\parbox{4.5cm}{ \scriptsize{\textbf{cmd1}: Go to the house on the left.}}}
   & \;TL $\downarrow$ & 8.54 & 8.60 & 8.40 & 8.52 \\
&  & \;SR $\uparrow$   & 61.01 & 92.19 & 95.32 & 91.82 \\
&  & \;NE $\downarrow$ & 2.80  & 3.11  & 2.40  & 2.40 \\
\cmidrule(lr){2-7}

& \multirow{3}{*}{\parbox{4.5cm}{ \scriptsize{\textbf{cmd2}: Go to the tree on the right.}}}
   & \;TL $\downarrow$ & 4.64 & 5.41 & 5.10 & 4.85 \\
&  & \;SR $\uparrow$   & 55.59 & 90.08 & 93.80 & 92.49 \\
&  & \;NE $\downarrow$ & 1.21  & 1.23  & 1.00  & 0.96 \\
\cmidrule(lr){2-7}

& \multirow{3}{*}{\parbox{4.5cm}{ \scriptsize{\textbf{cmd3}: Find the mailbox, and fly towards it.}}}
   & \;TL $\downarrow$ & 5.24 & 6.17 & 5.98 & 6.27 \\
&  & \;SR $\uparrow$   & 55.32 & 91.18 & 94.21 & 91.47 \\
&  & \;NE $\downarrow$ & 2.01  & 2.01  & 1.83  & 1.83 \\
\cmidrule(lr){2-7}

& \multirow{3}{*}{\parbox{4.5cm}{ \scriptsize{\textbf{cmd4}: Fly between the two houses,\\ look for a tree, and fly towards it.}}}
   & \;TL $\downarrow$ & 14.51 & 14.70 & 13.90 & 18.82 \\
&  & \;SR $\uparrow$   & 50.81 & 91.41 & 96.02 & 93.03 \\
&  & \;NE $\downarrow$ & 0.51  & 0.50  & 0.35  & 0.36 \\

\midrule

\multirow{12}{*}{\textbf{\small{city}}}

& \multirow{3}{*}{\parbox{7.5cm}{ \scriptsize{\textbf{cmd1}: Go past the first traffic light and go straight past the blue car.\\
After crossing a blue mailbox, turn right at the stop sign, and land in front of the gas station.}}}
   & \;TL $\downarrow$ & 86.37 & 89.50 & 89.30 & 87.12 \\
&  & \;SR $\uparrow$   & 58.32 & 88.90 & 94.45 & 90.10 \\
&  & \;NE $\downarrow$ & 3.20  & 3.40  & 2.90  & 3.10 \\
\cmidrule(lr){2-7}

& \multirow{3}{*}{\parbox{7.5cm}{ \scriptsize{\textbf{cmd2}: Follow the road. After the crossing, fly through the alley before the blue mailbox. Turn left, pass between buildings, turn left at the oak tree, and if a white truck is visible, land in front of it.}}}
   & \;TL $\downarrow$ & 118.60 & 129.19 & 112.83 & 126.15 \\
&  & \;SR $\uparrow$   & 52.25 & 87.30 & 93.21 & 89.99 \\
&  & \;NE $\downarrow$ & 1.60  & 1.50  & 1.25  & 1.30 \\
\cmidrule(lr){2-7}

& \multirow{3}{*}{\parbox{7.5cm}{ \scriptsize{\textbf{cmd3}: Head past the second traffic light. If an ambulance is on the left, fly past it and the stop sign. Enter the gazebo and land inside.}}}
   & \;TL $\downarrow$ & 169.80 & 179.70 & 172.29 & 174.80 \\
&  & \;SR $\uparrow$   & 54.10 & 89.50 & 94.12 & 90.55 \\
&  & \;NE $\downarrow$ & 2.40  & 2.10  & 1.80  & 1.95 \\
\cmidrule(lr){2-7}

& \multirow{3}{*}{\parbox{7.5cm}{ \scriptsize{\textbf{cmd4}: Fly past the first traffic light, then turn right before the gas station. Before the white truck, turn left. At the apartment with stairs, ascend to the third floor and land inside the hallway.}}}
   & \;TL $\downarrow$ & 139.10 & 143.70 & 140.02 & 147.39 \\
&  & \;SR $\uparrow$   & 48.92 & 89.60 & 95.42 & 91.80 \\
&  & \;NE $\downarrow$ & 0.70  & 0.60  & 0.50  & 0.55 \\

\bottomrule
\end{tabular}
}
\end{table*}

\begin{figure}[t]
    \centering
    \includegraphics[height = 6.2cm]{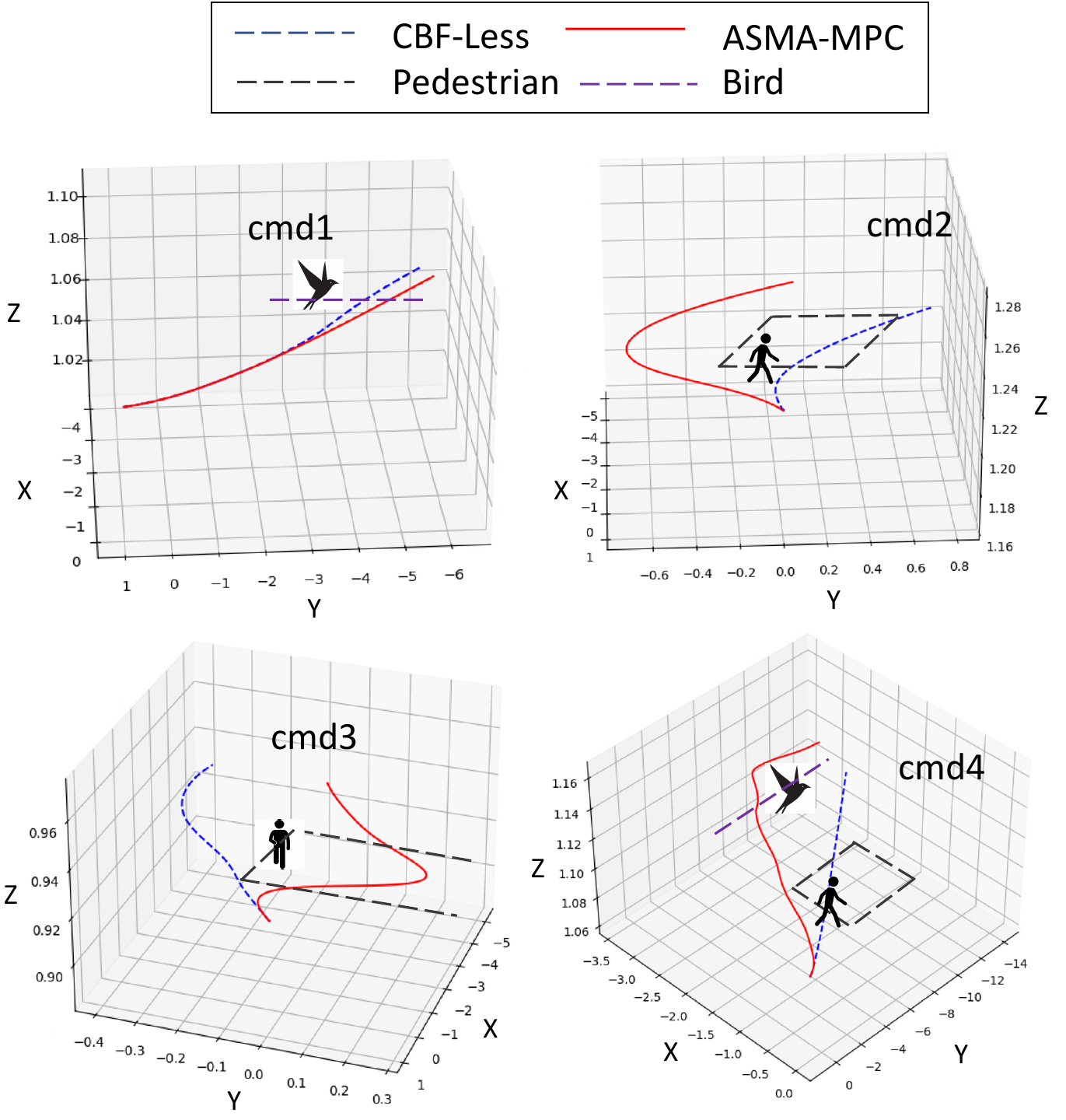}
    \caption{{Comparison of navigation trajectories for four VLN instructions in the \mbox{\textbf{small world}} environment. The blue dashed line represents the \mbox{\textbf{CBF-less}} method, while the red solid line corresponds to \mbox{\textbf{ASMA-MPC}}. Gray and purple dashed lines denote the trajectories of dynamic pedestrians and birds, respectively. Ghost icons indicate obstacle positions at the moment of closest encounter. ASMA-MPC demonstrates obstacle-aware local trajectory modulation in response to dynamic threats. }}
    \label{fig:traj_small_world}
\end{figure}

\begin{figure}[t]
    \centering
    \includegraphics[height=6.2cm]{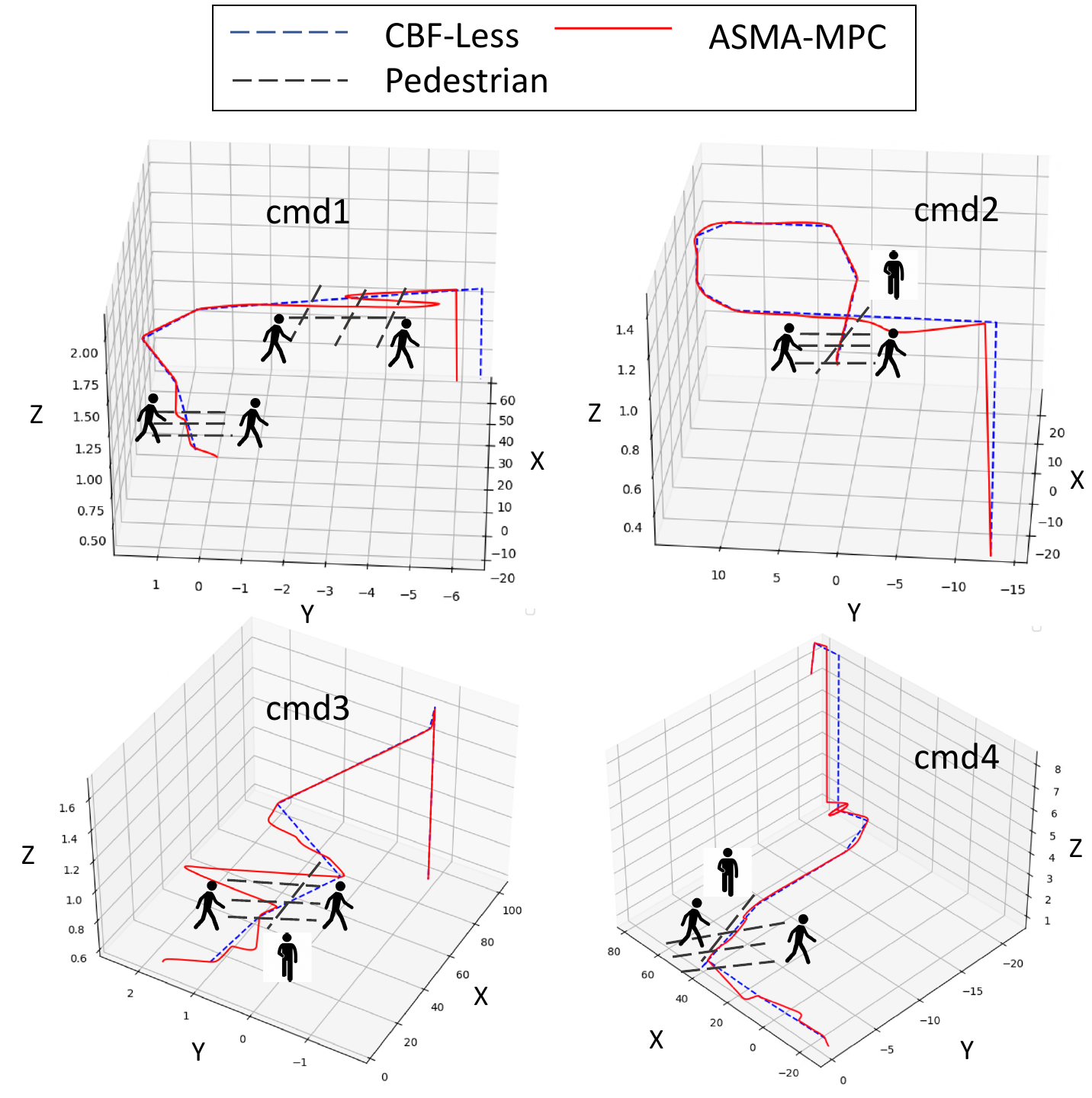}
    \caption{{Comparison of navigation trajectories for four VLN instructions in the \textbf{city} environment. The blue dashed line represents the \textbf{CBF-less} method, while the red solid line corresponds to \textbf{ASMA-MPC}. Black dashed curves represent pedestrian trajectories. Ghost pedestrian icons highlight positions at key avoidance moments. ASMA-MPC demonstrates smoother and safer trajectory adaptations in response to dense, dynamic pedestrian flows.}}
    \label{fig:traj_city}
    \vspace{-4mm}
\end{figure}


\begin{figure*}[!t]
    \centering
    \includegraphics[height=3cm]{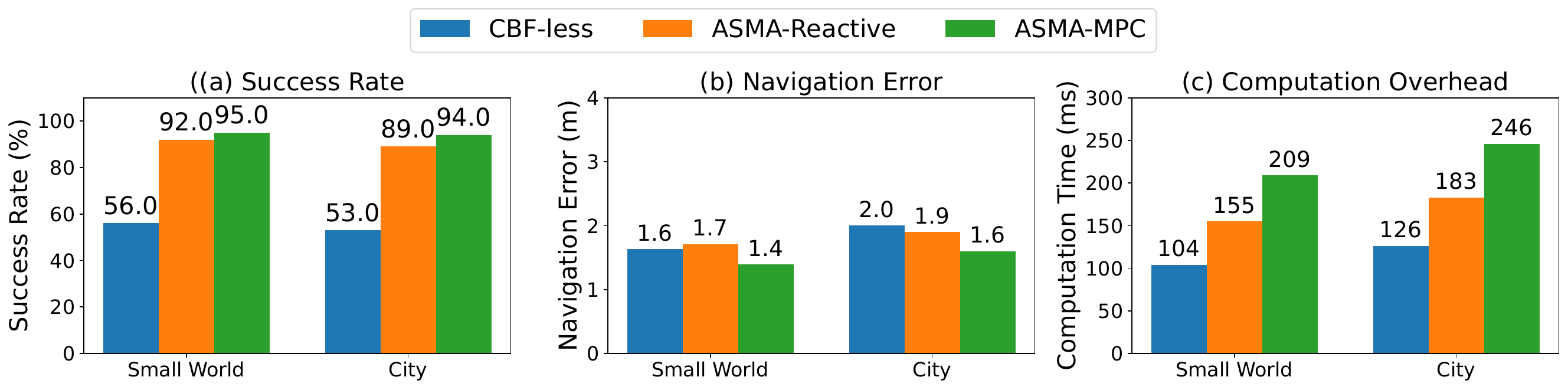}
    \vspace{-4mm}
    \caption{Ablation study results comparing the three methods 
    (CBF-less, ASMA-Reactive, and ASMA-MPC) across two environments 
    (Small World and City). Bars indicate average Success Rate (SR), 
    Navigation Error (NE), and Computation Time for 1 inference cycle.}
    \label{fig:ablation}
    \vspace{-6mm}
\end{figure*}
\vspace{-2mm}
\subsection{\textbf{Vision--Language Encoder Performance}}
We evaluate our vision–language encoder by computing cosine similarity between CLIP-encoded text queries and cropped object embeddings. This includes both direct detections and fallback region proposals. Grounding accuracy is defined as the percentage of correctly matched landmarks.
Table~\ref{tab:vl_performance} reports results for the \textit{small\_world} and \textit{city} environments. All similarity scores are obtained in a zero-shot setting, without fine-tuning, as the encoder combines outputs from a vision-language model and an object detector. While fine-tuning could improve scores, our goal is robust landmark extraction—fine-tuning may reduce generalization. This confirms that the encoder effectively grounds language to visual landmarks for downstream safety-aware control.

\vspace{-2mm}
\subsection{\textbf{Vision-Language Navigation Performance}}
Table~\ref{tab:vln} compares four methods in \textit{small\_world} (top) and \textit{city} (bottom). In \textit{small\_world}, CBF-less has low Success Rates (SR), while ASMA variants improve SR significantly (up to 96.02\%). ASMA-Reactive sometimes increases Trajectory Length (TL), whereas ASMA-MPC achieves a better SR–TL tradeoff. CBF-Only improves over baseline but lacks language grounding. In the more complex \textit{city}, ASMA-MPC anticipates hazards, reaching 95.42\% SR and low Navigation Error (NE). Figures~\ref{fig:traj_small_world} and~\ref{fig:traj_city} show that CBF-less (dashed blue) fails near obstacles, while ASMA-MPC (solid red) maintains safe, instruction-following paths. ASMA-Reactive and ASMA-MPC boost SR by 64.1\% and 67.5\%, with only 1.4\%–5.8\% TL increases from safety detours.

\subsection{\textbf{Ablation Study}}

\label{sec:ablation}
{Our ablation study compares three variants—CBF-less, ASMA-Reactive, and ASMA-MPC—across both \textit{small world} and \textit{city} environments. As shown in Figure~\mbox{\ref{fig:ablation}}, CBF-less yields lower success rates, while ASMA variants significantly improve performance. ASMA-MPC achieves the highest success rate and lowest navigation error by proactively enforcing safety via model predictive planning, at the cost of increased computation. To support latency-sensitive deployments, ASMA-Reactive offers a lower-latency alternative that enforces safety reactively. As seen in Figure~\mbox{\ref{fig:ablation}}(c), ASMA-MPC incurs higher latency but delivers better safety and task performance. Dynamic/static object classification is performed via lightweight bounding box comparisons and has negligible runtime cost. ASMA-MPC increases processing time by ~30–40\%, which is justified by its gains. All variants meet real-time constraints on our evaluation setup (Intel Core i9, RTX 3090, 2~GB). Further optimization is needed for embedded deployment. This highlights a practical tradeoff: ASMA-MPC for higher compute budgets, ASMA-Reactive for time-critical settings.}

\section{Discussion}
\label{sec:disc}

{The current framework advances state-of-the-art agentic navigation by integrating scene-aware Control Barrier Functions (CBFs) with CLIP-YOLO for language-conditioned visual navigation, enabling robust safety enforcement even in the presence of arbitrarily moving obstacles and complex structures, while effectively handling edge cases like occlusion or out-of-view objects. If a referred object is initially occluded or out of view, it is not detected immediately. However, the navigation policy proceeds using available spatial cues, and detection resumes once the object becomes visible. This enables implicit handling of partial observability, though the system does not explicitly model occlusion or object permanence. While not active in the current version, basic user feedback (e.g., “The object does not exist”) can be incorporated using a lightweight decoder trained on detection confidence and instruction progress. Mechanisms like visual memory, semantic mapping, or uncertainty-aware exploration may further improve handling of occlusions and ambiguity. ASMA supports extension to other perception modalities (e.g., segmentation, optical flow), making it a general framework for scene-aware CBFs in VLN safety. Real-world deployment with dynamic obstacles such as pedestrians or birds remains challenging, as aerial platforms raise safety and regulatory concerns which are better addressed in simulation. }

\section{Summary}

We discussed contemporary VLN methods do not address dynamic obstacles in evolving environments. We introduced ASMA (Adaptive Safety Margin Algorithm) to enhance VLN safety for drones using a novel scene-aware CBF formulation. By continuously identifying potentially risky
observations, the system performs prediction in real time about
unsafe conditions. ASMA dynamically adjusts control actions based on real-time depth data, ensuring safe navigation in complex environments. Compared to a baseline CBF-less VLN model, two variants of ASMA achieves 64\%--67\% increase in success rates with only a slight (1.4\%--5.8\%) increase in trajectory lengths compared to the baseline CBF-less VLN. 

\vspace{-1mm}

\bibliographystyle{IEEEtran}
\bibliography{conference_101719}

\end{document}